\documentclass[letterpaper]{article} 
\usepackage{aaai2027}  
\usepackage[hyphens]{url}  
\usepackage{graphicx} 
\usepackage{natbib}  
\usepackage{caption} 
\usepackage{algorithm}
\usepackage{algorithmic}
\usepackage{booktabs}
\usepackage{subcaption}
\usepackage{multirow}
\usepackage{amsmath}
\usepackage{makecell}
\usepackage{pifont}
\usepackage{xcolor,colortbl}
\definecolor{stableblue}{RGB}{225, 240, 250}
\definecolor{moderateyellow}{RGB}{255, 248, 220}
\definecolor{strongred}{RGB}{252, 228, 228}
\newcommand{\cmark}{\ding{51}}
\usepackage{newfloat}
\usepackage{listings}
\DeclareCaptionStyle{ruled}{labelfont=normalfont,labelsep=colon,strut=off} 
\floatstyle{ruled}
\newfloat{listing}{tb}{lst}{}
\floatname{listing}{Listing}

\usepackage{booktabs}

\title{Large Emotional World Model}
\author{
    Changhao Song\textsuperscript,
    Yazhou Zhang\textsuperscript\corresponding,
    Hui Gao\textsuperscript,
    Chang Yang\textsuperscript,
    Peng Zhang\textsuperscript\corresponding
}

\affiliations{
    Tianjin University, Tianjin, China\\
    songchanghao@tju.edu.cn,
    yzhou\_zhang@tju.edu.cn,
    hui\_gao@tju.edu.cn,\\
    yangchang@tju.edu.cn,
    pzhang@tju.edu.cn
}

\begin{document}

\maketitle

\begin{abstract}
The world is governed by both physical laws and affective dynamics. 
Physical laws govern state transitions, while affective dynamics shape human actions, decisions, and interactions.
A world model that learns only physical laws can approximate the physical world, but not the human world.
In this paper, we introduce human emotion as a key state variable in world models, enabling them to capture both future state transitions and their emotional causes.
We first construct \textbf{E}motion-\textbf{Why}-\textbf{How} (EWH), the first world model dataset centered on emotional state transitions, containing 10,850 emotion-aware transition tuples. Each tuple encodes the pre-state, pre-emotion, action, post-emotion, and post-state, supporting reasoning about why actions occur and how emotions reshape future states.
Based on EWH, we propose the \textbf{L}arge \textbf{E}motional \textbf{W}orld \textbf{M}odel (LEWM), which factorizes future prediction into two coupled steps: first predicting the future emotional state from the current context, and then conditioning future world-state prediction on the predicted emotion.
Experiments show that LEWM brings consistent gains across world-state prediction, emotion understanding, and general reasoning tasks.
It achieves up to 45.72\% accuracy improvement on EWH, 3.94\% on WorldNet, 17.47\% F1 improvement on MELD, and a 6.10\% gain on specific MMLU categories. These results demonstrate that incorporating emotion into world models enables more realistic simulation of human-centered environments and expands the predictive understanding of intelligent agents.
\footnote{The dataset is available upon request from the authors.}
\end{abstract}


\section{Introduction}
World models aim to learn an internal representation of the world and use it to predict how the world will evolve~\cite{ha2018world}. 
Instead of directly reacting to observed inputs, a world model builds a predictive simulator that captures the regularities underlying state transitions~\cite{hafner2025mastering}. This ability to model the future has made world models a central paradigm in machine intelligence, with broad applications in visual prediction, embodied learning, decision-making, and long-horizon reasoning~\cite{micheli2022transformers, ding2025understanding}. 

\begin{figure}[h]
    \centering
    \includegraphics[width=1.0\columnwidth]{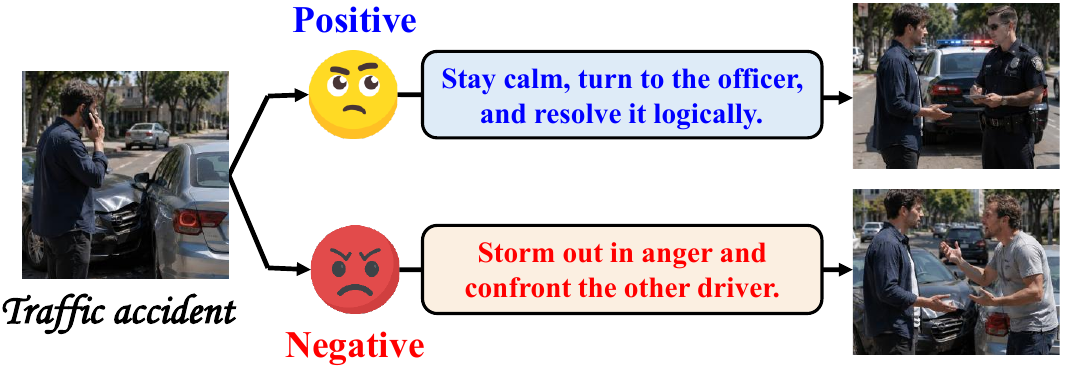}
    \caption{The same physical event leads to different future trajectories under different emotional states.}
    \label{fig:motivation}
\end{figure}

Before modeling the world, a world model must first confront a fundamental question: what kind of world is being modeled?
Prior studies suggest that the world is not merely a physical system governed by external laws, but also a human world shaped by affective dynamics~\cite{zhang2026world,searle1995construction}.
Physical laws explain how external states evolve, such as object motion, spatial change, and visual continuity. In contrast, human actions, decisions, and interactions are deeply shaped by internal affective states, including emotion, intention, etc.~\cite{lerner2015emotion, song2025emotion}. 
This distinction implies that a complete world model for human-centered environments should not only predict physical state evolution, but also model the affective causes that drive human actions and future changes.


Existing world models predict future states either directly in the observable space~\cite{brooks2024video,bruce2024genie,hu2023gaia,agarwal2025cosmos} or through latent transition dynamics~\cite{hafner2023mastering,assran2023self}. Despite their differences in architecture, both lines aim to capture regularities from current states to future states. However, current world models largely remain models of the physical world, but rarely treat emotion as a causal factor in world modeling. This omission would become critical in human-centered environments, where the same physical state can lead to different future transitions under different emotional states.
As shown in Fig.~\ref{fig:motivation}, following a traffic accident, positive and negative emotions influence whether the driver adopts cooperative or hostile behaviors, leading to effective or disrupted resolution.
Recent studies on world modeling have begun to emphasize that a complete model of the world should place the inference of human mental states alongside the modeling of physical dynamics~\cite{fung2025embodied}.
Therefore, a world model for human-centered environments must move beyond physical state prediction and incorporate affective dynamics into its core transition structure.

To fill this gap, we introduce emotional state transitions into world modeling. We first construct \textbf{E}motion-\textbf{Why}-\textbf{How} (EWH), the first world model dataset centered on emotional state transitions. EWH contains 10,850 emotion-aware transition tuples, each containing a pre-state, a pre-emotion, an action, a post-emotion, and a post-state. This design explicitly connects observable state changes with internal affective changes, enabling models to reason about why an action occurs and how emotion reshapes future states. 
Based on EWH, we further propose the \textbf{L}arge \textbf{E}motional \textbf{W}orld \textbf{M}odel (LEWM). The central idea of LEWM is to factorize future prediction into two coupled steps. First, the model predicts the future emotional state from the current context. Second, the predicted emotion is used as a conditioning signal to guide future world-state prediction. This design reflects the assumption that, in human-centered environments, future states are not determined by external observations and actions alone, but are mediated by affective transitions. 

We evaluate LEWM across three levels of capability. First, on world-state prediction tasks, we test whether emotion-aware modeling improves the prediction of future states in both emotion-centered and physical-world settings. Second, on emotion understanding tasks, we examine whether learning emotional state transitions enhances the model's ability to recognize and reason about human emotion. Third, on general reasoning benchmarks, we investigate whether incorporating emotional dynamics harms or preserves broader reasoning ability. Experimental results show that LEWM consistently improves performance across these settings, achieving up to \textbf{45.72\%} accuracy improvement on EWH, \textbf{3.94\%} on WorldNet, \textbf{17.47\%} F1 improvement on MELD, and a \textbf{6.10\%} gain on specific MMLU categories. These results demonstrate the effectiveness and generality of emotional world modeling.
Our contributions are three-fold:
\begin{itemize}
\item We formulate affective world modeling by introducing emotion into world-state transition modeling.
\item We build EWH, the first emotion-centered world model dataset with 10,850 transition tuples.
\item We propose LEWM and demonstrate its effectiveness across world-state prediction, emotion understanding, and general reasoning tasks.
\end{itemize}

\section{Related Work}

\subsubsection{World Model.}
World models learn predictive representations of state transitions to support prediction, planning, and policy learning~\cite{ha2018world,sutton1991dyna}. With the rise of autoregressive models, they have evolved from compact latent transition models into large-scale predictive models~\cite{ding2025understanding}. Existing approaches fall into two categories. Observable-space models generate future frames, scenes, or action-conditioned trajectories, as exemplified by Sora~\cite{brooks2024video}, Genie~\cite{bruce2024genie}, GAIA-1~\cite{hu2023gaia}, and Cosmos~\cite{agarwal2025cosmos}. In contrast, latent-state models compress observations into hidden representations and predict their transitions, including recurrent state-space and latent dynamics models such as Dreamer~\cite{hafner2023mastering} and JEPA~\cite{assran2023self}. By abstracting away pixel-level details, latent models are suitable for long-horizon prediction and reasoning.

Nevertheless, both lines rarely treat emotion as an explicit state variable that participates in the transition process. Our work extends world modeling by introducing affective dynamics into future prediction, enabling models to capture not only what changes in the world, but also how emotional states mediate these changes.

\begin{figure*}[h]
    \centering
    \includegraphics[width=1.0\textwidth]{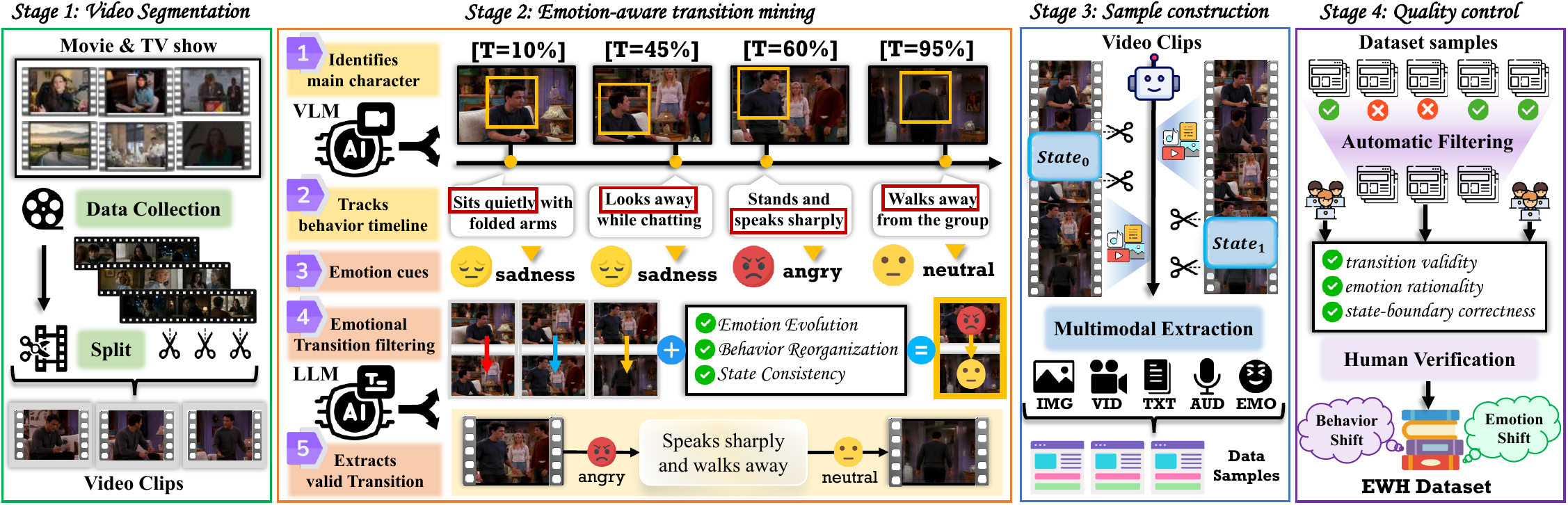}
    \caption{EWH Dataset Construction Pipeline.}
    \label{fig:dataset_pipeline}
\end{figure*}

\subsubsection{Emotion Understanding in Language Models.}
Recent LLM-based research~\cite{Xu2026Survey} follows two main directions. The first adapts LLMs to tasks such as sentiment analysis, emotion recognition, and sarcasm detection, with representative models including DialogueLLM~\cite{zhang2025dialoguellm}, and Emotion-LLaMA~\cite{xing2024emo}. The second investigates emotional reasoning. Early methods, such as ECR-Chain~\cite{huang2024ecr} and DECC~\cite{wu2024enhancing}, use chain-of-thought prompting to infer emotional causes, contextual cues, and affective states. SarcasmCue~\cite{yao2025sarcasm} further explores sequential and non-sequential reasoning structures for sarcasm detection. More recently, Emotion-o1~\cite{song2025emotion} and AffectGPT-R1~\cite{lian2025affectgpt} employ reinforcement learning for long-form emotional reasoning.

However, existing studies are still mostly framed as understanding or explanation tasks. They fail to connect emotional states with future world-state transitions in the sense of world modeling. In contrast, our work moves emotion modeling from static classification to dynamic emotional evolution.

\section{Emotion-Why-How(EWH) Dataset}
\label{subsec:emotional_world_model}


\subsection{Notation Definition}
\label{sec:problem_formulation}

To formalize EWH, we define each sample as an emotion-aware state transition. For $i\in\{0,1\}$, the observable state is
$\mathcal{S}_i=(\mathcal{V}_i,\mathcal{A}_i,\mathcal{I}_i)$,
where $\mathcal{V}_i$, $\mathcal{A}_i$, and $\mathcal{I}_i$ denote the synchronized video, audio, and image observations, respectively.
Let $\mathcal{E}_i$ denote the corresponding emotional state and $\mathcal{H}$ the observable behavior trajectory connecting the two states. Each EWH sample is represented as
\begin{equation}
\mathcal{T}
=
\left\langle
(\mathcal{S}_0,\mathcal{E}_0),
\mathcal{H},
(\mathcal{S}_1,\mathcal{E}_1)
\right\rangle
\end{equation}
where subscripts $0$ and $1$ denote the initial and future states of a transition sample, respectively.

\subsection{Dataset Construction} 
\label{sec:dataset} 

To emphasize real-world affective and behavioral dynamics, we first collect 2,258 first-person videos with emotional or behavioral turning points from \emph{TikTok} and \emph{Douyin}. We further collect videos from eight television series: \emph{Desperate Housewives}, \emph{Frasier}, \emph{Friends}, \emph{How I Met Your Mother}, \emph{Modern Family}, \emph{The Big Bang Theory}, \emph{The Office}, and \emph{This Is Us}. Selected for their rich interpersonal interactions and affective behaviors, these series contribute 1,583 episodes and over 630 hours of video across domestic, workplace, educational, dining, travel, medical, and public settings.

As illustrated in Fig.~\ref{fig:dataset_pipeline}, the pipeline consists of four stages: (1) \textit{Video Segmentation}, (2) \textit{Emotion-Aware Transition Mining}, (3) \textit{Sample Construction}, and (4) \textit{Quality Control}.

\textbf{Video segmentation.}
We detect shot boundaries and merge adjacent shots into semantically coherent candidate clips of 5--15s, aiming to preserve complete emotion-driven transitions. An additional 3s context is retained on both sides during annotation and quality control to reduce ambiguity.

\textbf{Emotion-aware transition mining.}
For each candidate clip, we employ a two-stage VLM--LLM pipeline. Qwen3.5-VL identifies the primary protagonist and generates a behavior timeline, inserting anchors $[T=t_i\%]$ at major observable changes in action, posture, movement, or interaction. All descriptions are grounded in the same protagonist, and clips lacking consistent subject tracking are rejected.

Conditioned on this timeline, GPT-5 infers the protagonist's emotion in each phase from the described facial and bodily cues, using seven discrete labels for consistent annotation and evaluation: \textit{neutral}, \textit{happiness}, \textit{sadness}, \textit{anger}, \textit{fear}, \textit{surprise}, and \textit{disgust}. We retain only transitions from a non-neutral emotion to a different emotion that involve sustained behavioral reorganization, such as changes in movement, task focus, interaction strategy, or engagement target. Transient expressions, minor postural changes, and routine, task-driven, or externally imposed behaviors are excluded. Each valid transition is represented as
$\langle(\mathcal{S}_0,\mathcal{E}_0),\mathcal{H},(\mathcal{S}_1,\mathcal{E}_1)\rangle$
with temporally grounded state intervals.

\textbf{Sample construction.}
We map the predicted temporal anchors to absolute video timestamps and extract the corresponding transition segments.
For each sample, we extract the video clip, audio track, key frames, initial and future state description, behavior-trajectory description, initial and future emotion labels. These components are organized into a unified emotion-aware world model sample. 

\textbf{Quality control.}
We ensure annotation reliability through automatic consistency filtering and human verification. Each candidate transition is mined five times and retained if the same emotion pair occurs in at least three runs. Three annotators further evaluate 10\% of the retained samples for transition validity, emotion plausibility, and boundary correctness. The corresponding Fleiss' $\kappa$ scores are 0.82, 0.76, and 0.65, indicating strong overall inter-annotator agreement. 

EWH is split at the episode level into training, validation, and test sets in an 8:1:1 ratio to prevent cross-split overlap.

\subsection{Dataset Analysis}
\label{sec:dataset_analysis}


Table~\ref{tab:dataset_statistics} summarizes the statistics of EWH. From 240,221 candidate segments, we retain 10,850 emotion-aware transitions. Each sample corresponds to a temporally grounded state transition, with average durations of 5.4,s and 7.0,s for $\mathcal{S}_0$ and $\mathcal{S}_1$. The average lengths of state and action descriptions are 21.5 and 29.7 words.

\begin{table}[h]
\centering
\resizebox{\columnwidth}{!}{
\begin{tabular}{cccccc}
\toprule
\textbf{Candidates} & \textbf{Selected} & \textbf{Avg. $S_0$ Dur.} & \textbf{Avg. $S_1$ Dur.} & \textbf{Avg. State Len.} & \textbf{Avg. Action Len.} \\
\midrule
240,221 & 10,850 & 5.4 s & 7.0 s & 21.5 words & 29.7 words\\
\bottomrule
\end{tabular}
}
\caption{Statistics of the EWH Dataset.}
\label{tab:dataset_statistics}
\end{table}

\begin{figure}[h]
\centering
\includegraphics[width=1.0\columnwidth]{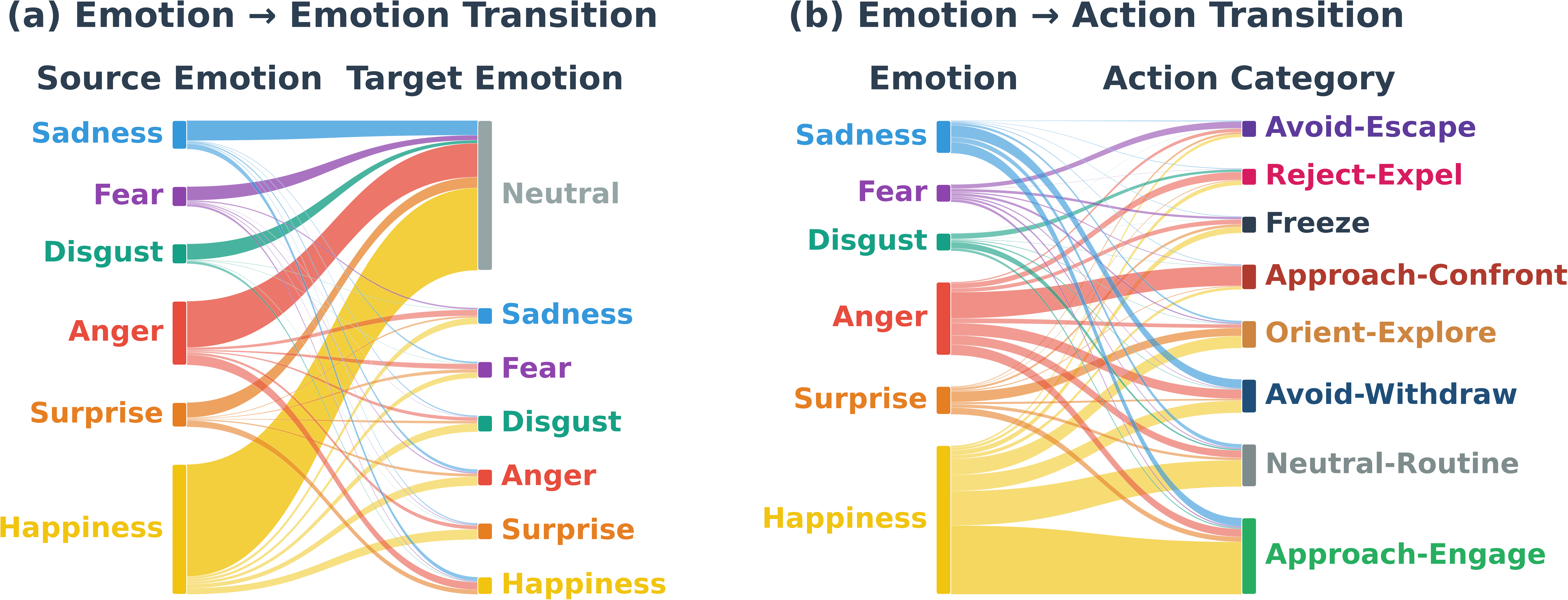}
\caption{Emotion transition and source distribution in EWH.}
\label{fig:statistics}
\end{figure}

\textbf{Emotion transition distribution.} Fig.~\ref{fig:statistics}(a) illustrates the emotion transition distribution in EWH. Nearly 80\% of emotional transitions evolve toward \textit{neutral}, indicating that EWH naturally captures emotion regulation and recovery processes rather than isolated emotional states. Meanwhile, \textit{happiness} and \textit{anger} account for over 74\% of source emotions, providing abundant samples of emotion-driven state evolution. Overall, these statistics highlight the dynamic nature of emotional transitions captured by EWH.

\textbf{Emotion-behavior association.} Inspired by the action tendency theory of emotion~\cite{fontaine2013emotion}, we categorize emotion-driven behaviors into eight representative action types. Fig.~\ref{fig:statistics}(b) reveals clear emotion-specific behavioral tendencies: positive emotions generally promote engagement, negative emotions are associated with confrontation or avoidance, whereas surprise more often leads to exploratory behaviors. These associations support the role of emotion as a causal driver of future world-state transitions.

\begin{figure}[h]
\centering
\includegraphics[width=0.9\columnwidth]{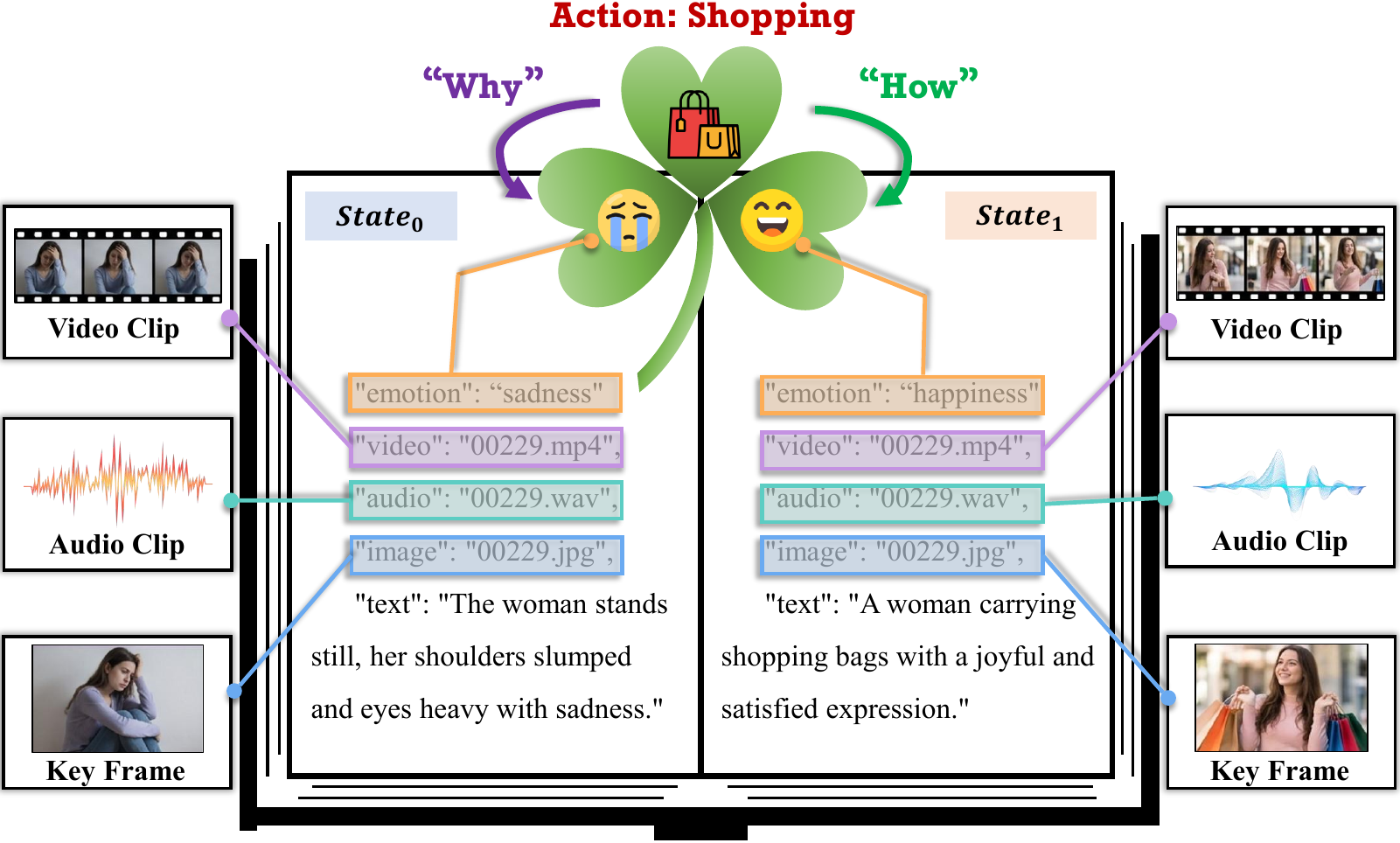}
\caption{Illustration of an EWH Dataset Sample.}
\label{fig:dataset}
\end{figure}

\textbf{Why-How Transition Structure.} As shown in Fig.~\ref{fig:dataset}, each sample follows a Why-How transition structure that connects the initial state-emotion pair, the intervening behavior, and the future state-emotion pair:
\begin{equation}
(\mathcal{S}_0,\mathcal{E}_0)
\xrightarrow{\text{Why}}
\mathcal{H}
\xrightarrow{\text{How}}
(\mathcal{S}_1,\mathcal{E}_1)
\label{eq:why_how}
\end{equation}

Here, \textit{Why} indicates the affective reason behind the observed behavior, while \textit{How} indicates the effect of emotional transition on future state evolution.

\subsection{Comparison with Existing Datasets}
\begin{table}[h]
\centering
\scriptsize
\resizebox{\columnwidth}{!}{
\begin{tabular}{lcccccc}
\toprule
Dataset & Text & Audio & Vision & Action & Emotion & Transition \\
\midrule
GoEmotions~\cite{demszky2020goemotions} & \cmark &        &        &        & \cmark &        \\
EMOTIC~\cite{kosti2019context}          &        &        & \cmark &        & \cmark &        \\
EmoSet~\cite{yang2023emoset}            &        &        & \cmark &        & \cmark &        \\
MELD~\cite{poria2018meld}               & \cmark & \cmark & \cmark &        & \cmark &        \\
CMMA~\cite{zhang2023cmma}      & \cmark &   \cmark     &    \cmark    &        & \cmark &        \\
CMU-MOSEI~\cite{zadeh2018multimodal}    & \cmark & \cmark & \cmark &        & \cmark &        \\
WorldNet~\cite{ge2024worldgpt}          & \cmark & \cmark & \cmark & \cmark &        & \cmark \\
\midrule
\textbf{EWH (Ours)}                     & \textbf{\cmark} & \textbf{\cmark} & \textbf{\cmark} & \textbf{\cmark} & \textbf{\cmark} & \textbf{\cmark} \\
\bottomrule
\end{tabular}
}
\caption{Comparison between EWH and Existing datasets.}
\label{tab:dataset_comparison}
\end{table}
As shown in Table~\ref{tab:dataset_comparison}, existing emotion datasets mainly provide emotional annotations for multimodel inputs, but they typically formulate emotion understanding as static recognition. In contrast, WorldNet introduces explicit state transitions but does not include emotional states. EWH bridges this gap by jointly representing multimodal observations, behavior, emotion, and future-state transitions.

\section{Large Emotional World Model (LEWM)}


\begin{figure*}[h]
    \centering
    \includegraphics[width=0.9\textwidth]{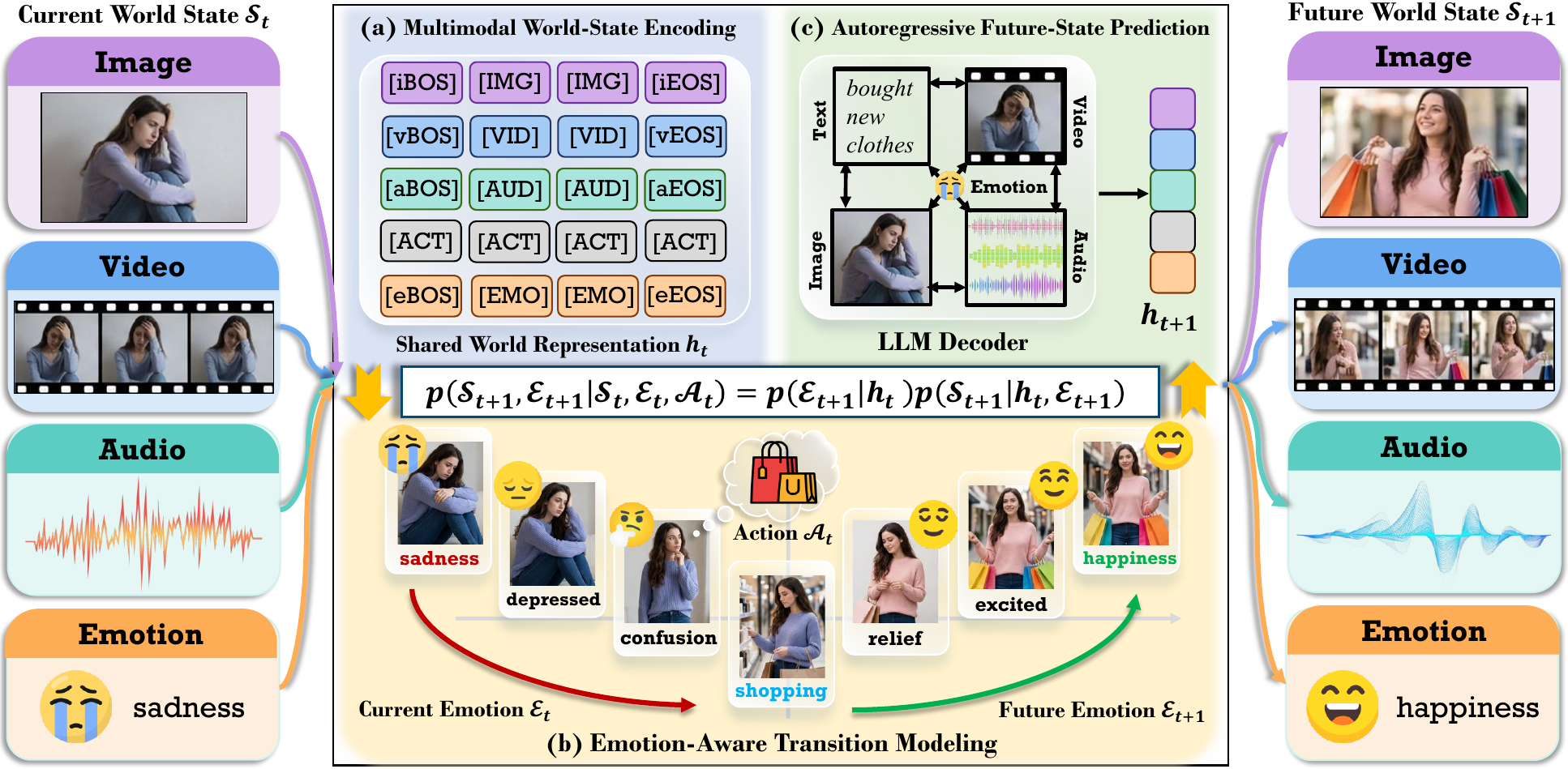}
    \caption{Overview of LEWM, a multimodal encoder--LLM--multimodal decoder framework for emotion-aware world modeling.}
    \label{fig:model}
\end{figure*}

\subsection{Problem Formulation}

Given the emotion-state pair $(\mathcal{S}_t,\mathcal{E}_t)$ at time $t$ and the intervening behavior trajectory $\mathcal{H}_t$, LEWM models the joint distribution of the multimodal state and emotion at $t+1$:

\begin{equation}
p(\mathcal{S}_{t+1},\mathcal{E}_{t+1}
\mid
\mathcal{S}_t,\mathcal{E}_t,\mathcal{H}_t)
\end{equation}
To explicitly model the role of emotion in future prediction, we factorize this transition into two coupled components:
\begin{equation}
\begin{aligned}
p(\mathcal{S}_{t+1},\mathcal{E}_{t+1}
\mid
\mathcal{S}_t,\mathcal{E}_t,\mathcal{H}_t) 
&=
p(\mathcal{E}_{t+1}
\mid
\mathcal{S}_t,\mathcal{E}_t,\mathcal{H}_t) \times\\
&\quad 
p(\mathcal{S}_{t+1}
\mid
\mathcal{S}_t,\mathcal{E}_t,\mathcal{H}_t,\mathcal{E}_{t+1})
\end{aligned}
\end{equation}
The first term models emotional transition, predicting how human's affective state evolves under the current context and behavior. The second term models emotion-conditioned world-state transition, predicting the future state based on both observable dynamics and the predicted emotional state.

\subsection{Overall Architecture}
As illustrated in Fig.~\ref{fig:model}, LEWM consists of three key components that operate sequentially:
(1) \textit{Multimodal World-State Encoding}, 
(2) \textit{Emotion-Aware Transition Modeling}, and
(3) \textit{Autoregressive Future-State Prediction}. 


\paragraph{Multimodal World-State Encoding.}
Given the observable world state
$\mathcal{S}_t=(\mathcal{V}_t,\mathcal{A}_t,\mathcal{I}_t)$
and the corresponding emotional state
$\mathcal{E}_t$,
LEWM first encodes each observable modality using a modality-specific encoder
$f_q(\cdot)$
followed by a trainable projection layer
$W_q$
to map modality features into the LLM embedding space:
\begin{equation}
u_t^q
=
W_qf_q(\mathcal X_t^q),
\qquad
q\in\{\mathcal V,\mathcal A,\mathcal I\}
\end{equation}
where
$\mathcal X_t^{\mathcal V}=\mathcal V_t$,
$\mathcal X_t^{\mathcal A}=\mathcal A_t$,
and
$\mathcal X_t^{\mathcal I}=\mathcal I_t$.
The emotional state
$\mathcal E_t$
is represented by a learnable emotion embedding table and projected into the same embedding space, yielding
$u_t^{\mathrm{emo}}$.
Here,
$W_q$
is a trainable linear projection matrix.
Collectively, the multimodal representations are denoted by
$\{u_t^q\}_{q\in\{\mathcal V,\mathcal A,\mathcal I\}}$
together with
$u_t^{\mathrm{emo}}$.

The behavior trajectory $\mathcal{H}_t$ is represented as a textual description of the intervening action process and encoded by the LLM tokenizer and embedding layer. We then wrap each modality and emotion representation with modality-specific boundary tokens to construct the input context:
\begin{equation}
X_t=
\left[
\{\tau_q^{\mathrm{B}},u_t^q,\tau_q^{\mathrm{E}}\}_{q\in\{\mathcal{V},\mathcal{A},\mathcal{I}\}};
\tau_{\mathrm{emo}}^{\mathrm{B}},
u_t^{\mathrm{emo}},
\tau_{\mathrm{emo}}^{\mathrm{E}};
\mathcal{H}_t
\right]
\end{equation}
where $\tau_q^{\mathrm{B}}$ and $\tau_q^{\mathrm{E}}$ denote the begin and end tokens for modality $q$, such as \texttt{<iBOS>}/\texttt{<iEOS>}, \texttt{<vBOS>}/\texttt{<vEOS>}, and \texttt{<aBOS>}/\texttt{<aEOS>}. Similarly, \texttt{<eBOS>} and \texttt{<eEOS>} indicate the boundary tokens for the emotional state.


\paragraph{Emotion-Aware Transition Modeling.}

Given the input context $X_t$, the LLM backbone performs autoregressive reasoning and produces a contextual representation
$z_t=\mathrm{LLM}(X_t)$,
where $z_t$ denotes the hidden representation used for transition prediction.
Following the emotion-first factorization defined in Eq.~(4), LEWM first predicts the future emotional state and then conditions future world-state prediction on the predicted emotion.

Specifically, the emotional transition head estimates the probability distribution over future emotional states:
\begin{equation}
p(\mathcal{E}_{t+1}
\mid
\mathcal{S}_t,\mathcal{E}_t,\mathcal{H}_t)
=
\mathrm{Softmax}(W_e z_t)
\end{equation}
where $W_e$ is a trainable emotion prediction head. 

The predicted future emotion is obtained by selecting the class with the highest probability:
\begin{equation}
\hat{\mathcal{E}}_{t+1}
=
\arg\max
p(\mathcal{E}_{t+1}
\mid
\mathcal{S}_t,\mathcal{E}_t,\mathcal{H}_t)
\end{equation}

The predicted emotion label
$\hat{\mathcal{E}}_{t+1}$
is then mapped through the shared emotion embedding table to obtain
$u_{t+1}^{\mathrm{emo}}$,
which is appended to the autoregressive context:
\begin{equation}
\tilde{X}_t=
\left[
X_t;
\tau_{\mathrm{emo}}^{\mathrm{B}},
u_{t+1}^{\mathrm{emo}},
\tau_{\mathrm{emo}}^{\mathrm{E}}
\right]
\end{equation}

Conditioned on $\tilde{X}_t$, LEWM predicts the future world state $\mathcal{S}_{t+1}$ through the multimodal decoder. This design enables LEWM to model how affective changes mediate human-centered world-state transitions.





\begin{table*}[h]
\centering
\small
\setlength{\tabcolsep}{3pt}
\resizebox{\textwidth}{!}{
\begin{tabular}{l|ccccc|ccccc}
\toprule
\multirow{2}{*}{\textbf{Task / Model}}
& \multicolumn{5}{c|}{\textbf{WorldNet (Physical World Dataset)}}
& \multicolumn{5}{c}{\textbf{EWH (Emotional World Dataset)}} \\
\cmidrule(lr){2-6}\cmidrule(lr){7-11}
& NExT-GPT & CoDi & WorldGPT & LEWM* & LEWM
& NExT-GPT & CoDi & WorldGPT & LEWM* & LEWM \\
\midrule

\multicolumn{11}{c}{\textbf{Single-Modal Prediction}} \\
\midrule

(image)$\rightarrow$(image)
& 0.3510 & 0.4969 & \textbf{0.5640} & \underline{0.5614} & 0.5503
& 0.3107 & 0.4382 & 0.4157 & \underline{0.5950} & \textbf{0.5967} \\

(video)$\rightarrow$(video)
& 0.4277 & 0.5351 & \underline{0.5918} & 0.5877 & \textbf{0.6066}
& 0.3489 & 0.5009 & 0.2232 & \underline{0.5572} & \textbf{0.6732} \\

(audio)$\rightarrow$(audio)
& 0.2552 & 0.3539 & 0.4086 & \underline{0.4437} & \textbf{0.4473}
& 0.2044 & \textbf{0.5431} & 0.2544 & 0.2974 & \underline{0.4949} \\

\midrule
\multicolumn{11}{c}{\textbf{Cross-Modal Prediction}} \\
\midrule

(audio, image)$\rightarrow$(video)
& 0.4245 & 0.5493 & 0.5820 & \underline{0.5823} & \textbf{0.6027}
& 0.3502 & 0.5108 & 0.2155 & \underline{0.5574} & \textbf{0.6727} \\

(audio, video)$\rightarrow$(image)
& 0.3463 & 0.4250 & \underline{0.5616} & \textbf{0.5637} & 0.5530
& 0.3121 & 0.3739 & 0.4303 & \underline{0.5951} & \textbf{0.5968} \\

(image)$\rightarrow$(video)
& 0.4269 & \underline{0.6035} & 0.5870 & 0.5835 & \textbf{0.6077}
& 0.3502 & 0.4286 & 0.2161 & \underline{0.5575} & \textbf{0.6727} \\

(image)$\rightarrow$(audio)
& 0.2552 & 0.2881 & 0.4072 & \underline{0.4430} & \textbf{0.4452}
& 0.2072 & 0.3453 & 0.2427 & \underline{0.2973} & \textbf{0.4902} \\

(video)$\rightarrow$(image)
& 0.3481 & 0.4212 & \underline{0.5650} & \textbf{0.5650} & \underline{0.5578}
& 0.3122 & 0.3661 & 0.4220 & \underline{0.5950} & \textbf{0.5966} \\

(video)$\rightarrow$(audio)
& 0.2571 & 0.3309 & 0.4072 & \underline{0.4432} & \textbf{0.4458}
& 0.2077 & \underline{0.4767} & 0.2484 & 0.2976 & \textbf{0.4885} \\

(audio, image, video)$\rightarrow$(image)
& 0.3472 & 0.4935 & \underline{0.5648} & \textbf{0.5661} & 0.5570
& 0.3093 & 0.4240 & 0.4222 & \underline{0.5948} & \textbf{0.5972} \\

(audio, image, video)$\rightarrow$(video)
& 0.4262 & 0.5796 & 0.5886 & \underline{0.5913} & \textbf{0.6073}
& 0.3505 & 0.5348 & 0.2150 & \underline{0.5570} & \textbf{0.6723} \\

(audio, image, video)$\rightarrow$(audio)
& 0.2532 & 0.3640 & 0.3935 & \textbf{0.4339} & \underline{0.4329}
& 0.2048 & \textbf{0.5717} & 0.2516 & 0.2968 & \underline{0.4898} \\

\bottomrule
\end{tabular}
}
\caption{Performance comparison on the Physical World Dataset (WorldNet) and Emotional World Dataset (EWH). The best and second-best results in each dataset are highlighted in bold and underlined, respectively.}
\label{tab:main_results}
\end{table*}

\paragraph{Autoregressive Future-State Prediction.}
To realize the above emotion-aware transition model, LEWM reformulates future-state prediction as modality-specific token generation by introducing dedicated generation tokens into the LLM vocabulary, including
$
[\mathrm{IMG}_i]
$,
$
[\mathrm{VID}_i]
$,
and
$
[\mathrm{AUD}_i]
$.
Conditioned on the augmented context
$
\tilde{X}_t
$,
the LLM autoregressively predicts the target generation-token sequence
$
Y_{t+1}
$:

\begin{equation}
p_{\theta}(Y_{t+1}\mid \tilde{X}_t)
=
\prod_{i=1}^{N}
p_{\theta}
(y_i
\mid
\tilde{X}_t,
y_{<i})
\end{equation}

The hidden states corresponding to the generated modality tokens are regarded as modality-specific transition representations. For each observable modality
$
q\in
\{\mathcal V,\mathcal A,\mathcal I\},
$
let
$
H_q
$
denote the hidden states at the corresponding generation-token positions. LEWM maps these representations through a modality-specific prediction head
$
\mathcal T_q
$:

\begin{equation}
\hat{\mathcal S}_{t+1}^{q}
=
\mathcal T_q(H_q),
\qquad
q\in
\{\mathcal V,\mathcal A,\mathcal I\}
\end{equation}

The predicted latent representations are subsequently mapped back to their original modality spaces through modality-specific decoders, yielding the future multimodal world state
$
\hat{\mathcal S}_{t+1}
=
(\hat{\mathcal V}_{t+1},
\hat{\mathcal A}_{t+1},
\hat{\mathcal I}_{t+1}).
$



To avoid over-reliance on predicted emotion, LEWM preserves a direct context-to-state pathway by combining $z_t$ and $u_{t+1}^{\mathrm{emo}}$ in the augmented context $\tilde{X}_t$. Future-state prediction therefore remains grounded in contextual dynamics when emotion prediction is imperfect, with emotion serving as a modulating signal rather than the sole determinant.

\subsection{Optimization}
LEWM is trained with a multi-task objective that jointly supervises future-emotion, autoregressive generation-token, and multimodal future-state prediction:

\begin{equation}
\mathcal{L}
=
\mathcal{L}_{\mathrm{LM}}
+
\alpha \mathcal{L}_{\mathrm{state}}
+
\lambda \mathcal{L}_{\mathrm{emotion}}
\end{equation}
where $\mathcal{L}_{\mathrm{LM}}$ is the standard autoregressive language modeling loss that supervises the generation of modality-specific target tokens, and
$\mathcal{L}_{\mathrm{emotion}}$
is implemented as a cross-entropy loss between the predicted future emotion
$\hat{\mathcal{E}}_{t+1}$
and the ground-truth emotion
$\mathcal{E}_{t+1}$.
The multimodal state loss
$\mathcal{L}_{\mathrm{state}}$
encourages the predicted future states to match the target states in the shared embedding space and is defined as

\begin{equation}
\mathcal{L}_{\mathrm{state}}
=
\frac{1}{|\mathcal{M}|}
\sum_{q\in\mathcal{M}}
d\!\left(
\hat{\mathcal{S}}_{t+1}^{q},
\mathcal{S}_{t+1}^{q}
\right)
\end{equation}
where
$d(\cdot,\cdot)$
denotes the embedding-space distance function and
$\mathcal{M}
=
\{\mathcal V,\mathcal A,\mathcal I\}$
is the set of observable modalities. The coefficients
$\alpha$
and
$\lambda$
control the relative importance of multimodal future-state prediction and emotion transition learning during optimization.

\section{Experiments}


\subsection{Experimental Settings}
\subsubsection{Datasets.}
We evaluate world-state prediction on the proposed Emotional World Dataset (EWH) and the physical-world benchmark WorldNet~\cite{ge2024worldgpt}; emotion understanding on MELD~\cite{poria2018meld}, CMMA~\cite{zhang2023cmma}, and GoEmotions~\cite{demszky2020goemotions}; and general reasoning on MMLU~\cite{hendryckstest2021}, HellaSwag~\cite{zellers2019hellaswag}, VQAv2~\cite{goyal2017making}, and NExT-QA~\cite{xiao2021next}. We further conduct ablation studies on EWH to examine key design choices.

\subsubsection{Baselines.}
We compare LEWM with CoDi~\cite{tang2023any} and NExT-GPT~\cite{wu2023next}, two representative any-to-any multimodal generation models, as well as WorldGPT, a representative world-modeling baseline. We further implement \textbf{LEWM*}, obtained by fine-tuning WorldGPT on proposed EWH dataset while keeping the original architecture unchanged, to distinguish the benefit of the dataset from that of our emotion-aware transition modeling. 

\subsection{Main Results}

Table~\ref{tab:main_results} reports cosine similarity for single- and cross-modal state prediction on Physical World Dataset WorldNet~\cite{ge2024worldgpt} and Emotional World Dataset EWH.


On EWH, LEWM achieves the best performance on 10 of 12 tasks, including all image and video tasks. Compared with WorldGPT, it improves video prediction by up to \textbf{45.73\%} and audio prediction by up to \textbf{24.75\%}, demonstrating the benefit of modeling emotional transitions in human-centered environments. Although CoDi performs better on 2 audio tasks, LEWM remains competitive across modalities.

On WorldNet, the improvements are smaller but remain consistent across tasks, with the largest gain reaching \textbf{3.94\%}. The largest gains occur in audio prediction, whereas image and video generation remain largely comparable to WorldGPT. This indicates that emotional dynamics provide complementary cues for modeling physical world transitions without compromising conventional state prediction.

LEWM* also surpasses WorldGPT on EWH, improving video, image, and audio generation on EWH by approximately \textbf{33.40\%}, \textbf{17.93\%}, and \textbf{4.30\%}, respectively, demonstrating the effectiveness of the proposed dataset. LEWM further improves video and audio generation by approximately \textbf{11.60\%} and \textbf{19.75\%}, respectively, showing the additional value of explicit emotion-aware transition modeling. In contrast, only marginal differences are observed on WorldNet, suggesting that emotion-aware transition modeling is particularly beneficial in human-centered environments.



\subsection{Extended Analysis}

\subsubsection{Emotion Understanding Capability.}

\begin{figure}[h]
    \centering
    \includegraphics[width=0.9\columnwidth]{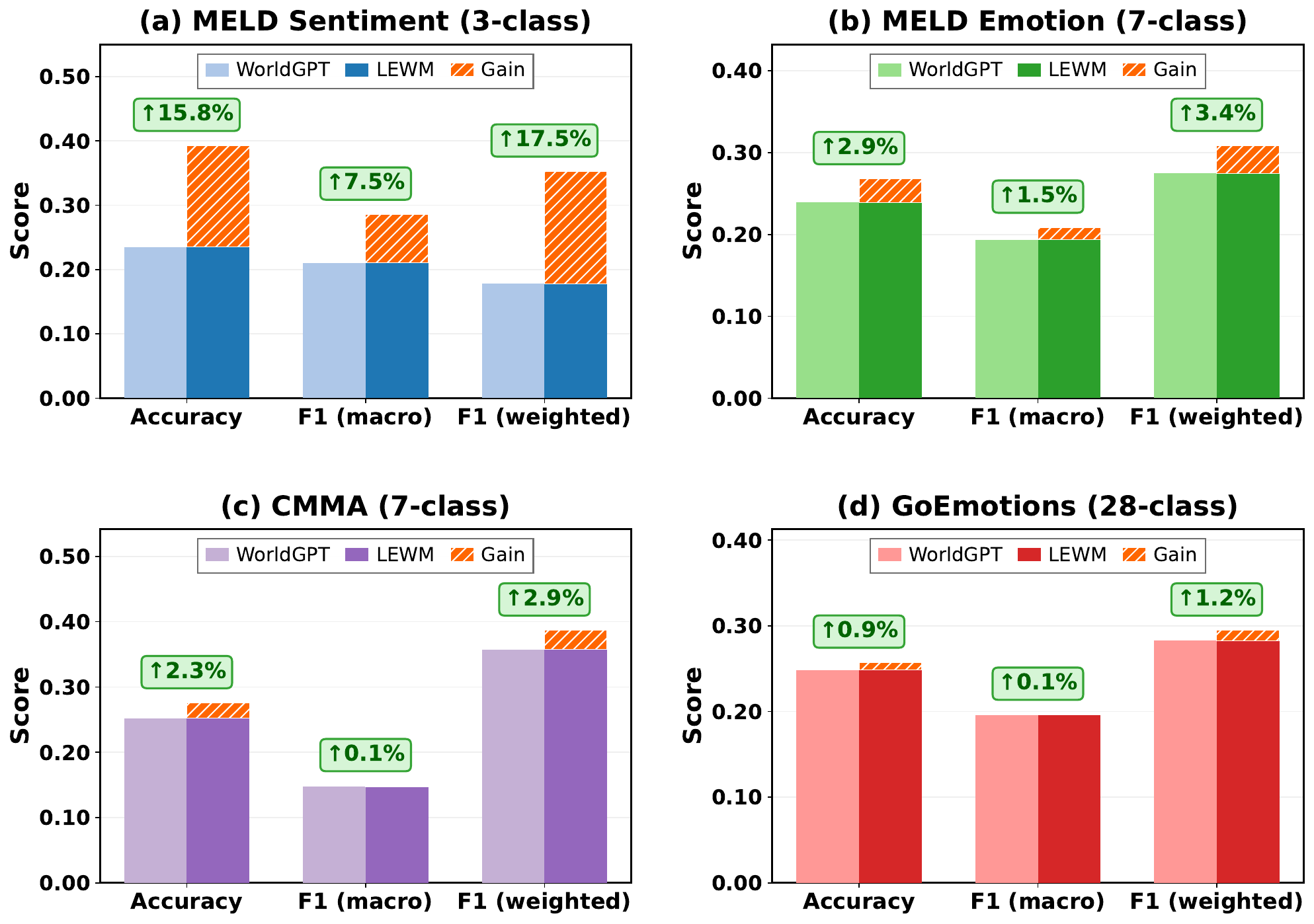}
    \caption{Comparison of emotion understanding performance on multi-modal and multi-class emotion benchmarks.}
    \label{fig:emotion_comparison}
\end{figure}

Figure~\ref{fig:emotion_comparison} shows that LEWM improves over WorldGPT across all datasets and metrics. The largest improvements are achieved on the MELD sentiment task, with absolute gains of \textbf{15.78\%} in Acc and \textbf{17.47\%} in Weighted-F1, while consistent improvements are also observed on the more fine-grained emotion benchmarks, including gains of \textbf{3.37\%} and \textbf{2.92\%} in Weighted-F1 on MELD and CMMA. And LEWM consistently outperforms its backbone even on the challenging 28-class GoEmotions benchmark, demonstrating that emotion-aware transition modeling learns transferable emotional representations.

\subsubsection{General Reasoning Compatibility.}


\begin{figure}[h]
    \centering
    \includegraphics[width=0.9\columnwidth]{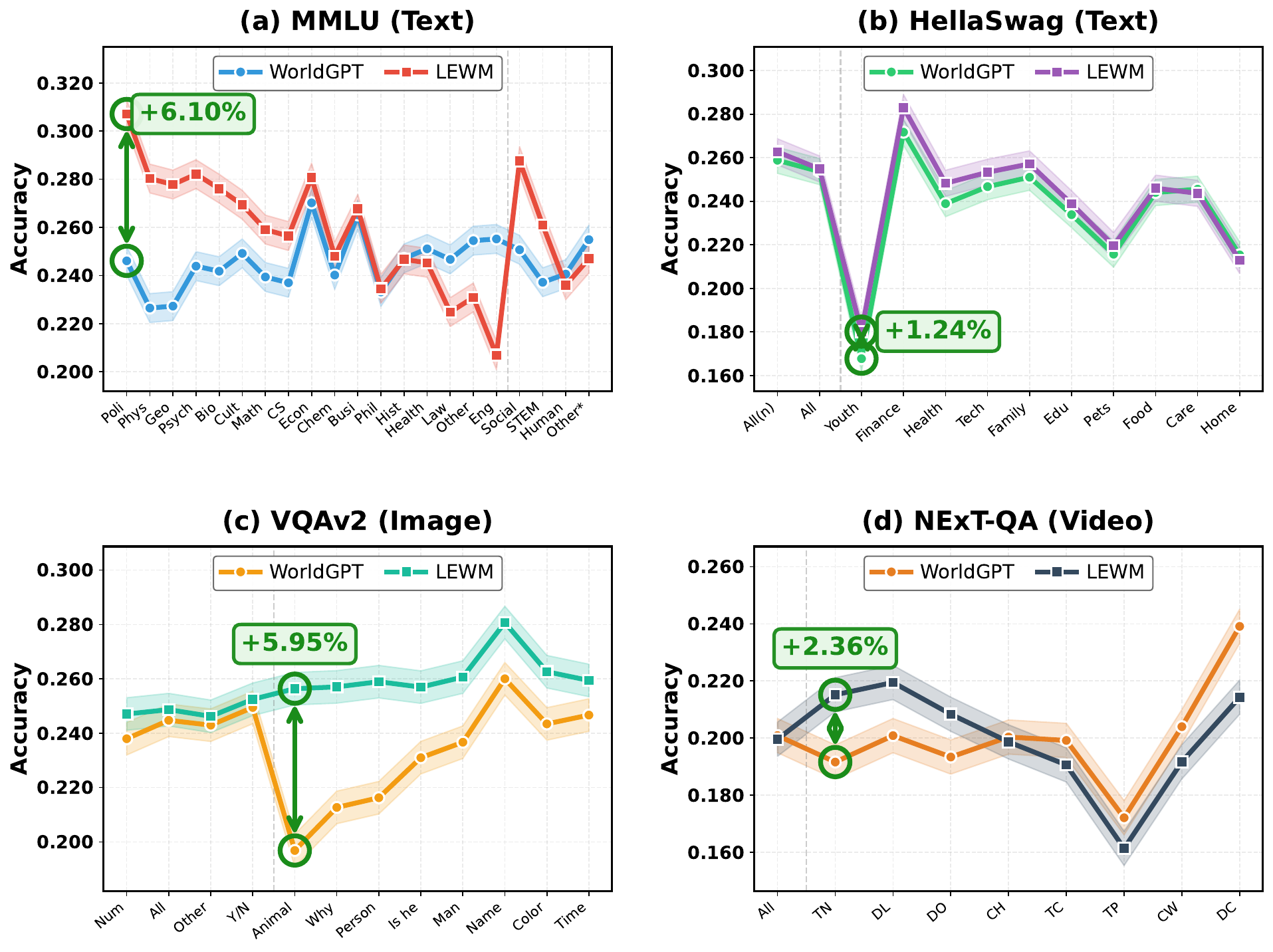}
    \caption{Comparison of general reasoning performance on text, image, and video benchmarks.}
    \label{fig:general_comparison}
\end{figure}

Figure~\ref{fig:general_comparison} shows that although performance fluctuates across individual categories, LEWM remains overall comparable to WorldGPT and even achieves notable gains on several tasks. The largest gains occur in \textit{Political Science} on MMLU and \textit{Animal} on VQAv2, reaching \textbf{6.10\%} and \textbf{5.95\%}, respectively, with smaller gains of up to \textbf{1.24\%} on HellaSwag and \textbf{2.36\%} on NExT-QA. These results suggest emotion-aware modeling and general reasoning are largely complementary, with certain reasoning domains even benefiting from explicit emotion modeling.


\subsubsection{Ablation Study on Emotion Conditioning.}


To assess emotion conditioning, we compare LEWM with five variants: \textbf{No Emotion} removes all emotional inputs; \textbf{Current Only} uses only the current emotion; \textbf{Hard Future} and \textbf{Soft Future} condition prediction on the most likely future-emotion label and its full distribution, respectively; and \textbf{Oracle Future} uses the ground-truth future emotion.

\begin{table}[h]
\centering
\scalebox{0.8}{
\begin{tabular}{lccc}
\toprule
\textbf{Variant} & \textbf{Image} & \textbf{Video} & \textbf{Audio} \\
\midrule
WorldGPT      & 0.4226 & 0.2175 & 0.2493 \\
\midrule
No Emotion    & 0.5950 & 0.5572 & 0.2974 \\
Current Only  & 0.5378 & 0.6649 & 0.2912 \\
Hard Future   & 0.5407 & 0.6606 & 0.2884 \\
Soft Future   & 0.5594 & 0.6593 & 0.3069 \\
Oracle Future & 0.4690 & \textbf{0.7515} & 0.4100 \\
\midrule
LEWM          & \textbf{0.5967} & 0.6732 & \textbf{0.4949} \\
\bottomrule
\end{tabular}
}
\caption{Ablation study on emotion conditioning strategies.}
\label{tab:emotion_ablation}
\end{table}

As shown in Table~\ref{tab:emotion_ablation}, removing emotion conditioning consistently degrades future-state prediction. \textit{Current Only} underperforms the full model, especially on audio (-20.37\%), supporting the value of modeling future emotional dynamics. Compared with \textit{Hard Future}, \textit{Soft Future} improves image and audio prediction by +1.87\% and +1.85\%, respectively, suggesting that preserving emotion uncertainty reduces error propagation. Finally, \textit{Oracle Future} outperforms \textit{Soft Future} on video and audio by +9.22\% and +10.31\%, indicating further potential from more accurate emotion prediction.

\subsubsection{Analysis by Emotion Transition Type.}

To further analyze the effect of emotion transitions, we group EWH samples 
by valence shifts: \textit{happiness} and \textit{joy} is positive, \textit{neutral} and \textit{surprise} are neutral, and the remaining emotions are negative.

\begin{table}[h]
\centering
\scalebox{0.8}{  
\renewcommand{\arraystretch}{1.0}
\setlength{\tabcolsep}{5pt}
\begin{tabular}{l|ccc}
\toprule
\textbf{Transition} & \textbf{Image} & \textbf{Video} & \textbf{Audio} \\
\midrule
\rowcolor{stableblue}
Neutral$\rightarrow$Neutral   & +18.77\% & \textbf{+47.02\%} & +23.42\% \\
\rowcolor{stableblue}
Negative$\rightarrow$Negative & +16.22\% & +44.83\% & +23.64\% \\
\midrule
\rowcolor{moderateyellow}
Positive$\rightarrow$Neutral  & +17.65\% & +46.08\% & +24.11\% \\
\rowcolor{moderateyellow}
Neutral$\rightarrow$Positive  & \textbf{+20.57\%} & +46.38\% & +24.02\% \\
\rowcolor{moderateyellow}
Neutral$\rightarrow$Negative  & +15.51\% & +46.17\% & +23.44\% \\
\rowcolor{moderateyellow}
Negative$\rightarrow$Neutral  & +17.14\% & +44.58\% & +22.22\% \\
\midrule
\rowcolor{strongred}
Positive$\rightarrow$Negative & +15.89\% & +40.65\% & +24.55\% \\
\rowcolor{strongred}
Negative$\rightarrow$Positive & +15.89\% & +44.30\% & \textbf{+25.11\%} \\
\bottomrule
\end{tabular}
}
\caption{Improvement (\%) of LEWM over WorldGPT,
grouped by emotion transition direction.
Rows are ordered by transition intensity:
\colorbox{stableblue}{Stable},
\colorbox{moderateyellow}{Moderate}, and
\colorbox{strongred}{Strong}.}
\label{tab:emotion_transition}
\end{table}

As shown in Table~\ref{tab:emotion_transition}, different modalities benefit differently from emotion-aware transition modeling. Specifically, the largest improvements are achieved for \textit{Neutral}$\rightarrow$\textit{Positive} in image prediction (+20.57\%), \textit{Neutral}$\rightarrow$\textit{Neutral} in video prediction (+47.02\%), and \textit{Negative}$\rightarrow$\textit{Positive} in audio prediction (+25.11\%). Overall, visual modalities benefit more from relatively continuous affective evolution, whereas audio shows greater gains under larger emotional shifts, suggesting that different modalities capture complementary aspects.

\subsubsection{Counterfactual Emotion Intervention.}

\begin{figure}[h]
    \centering
    \includegraphics[width=1.0\columnwidth]{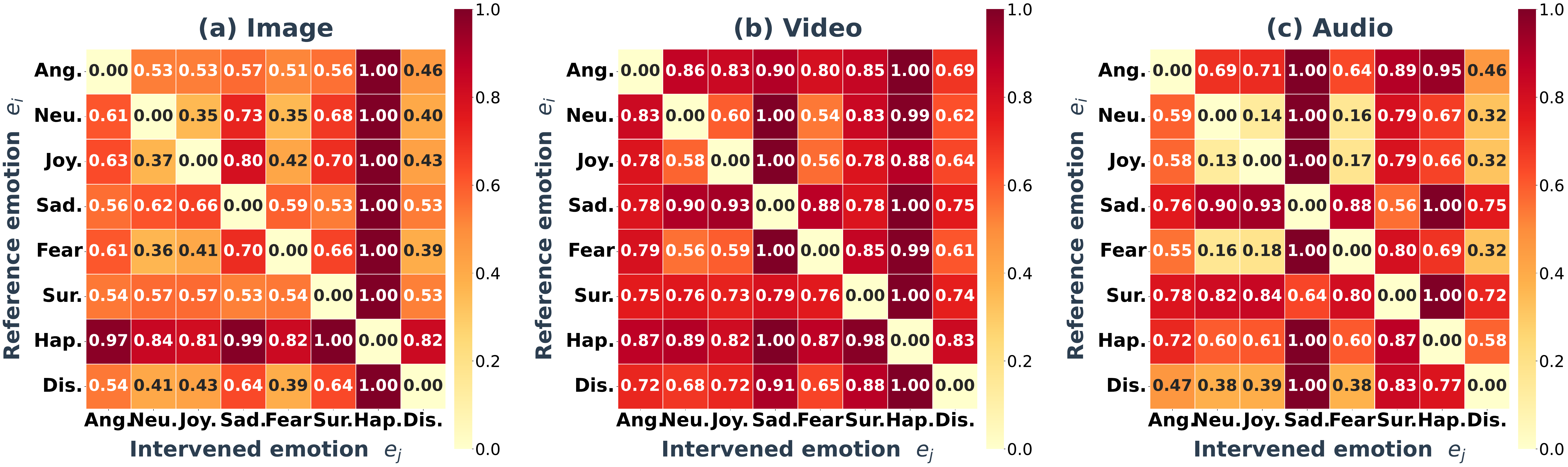}
    \caption{Counterfactual divergence under Emo intervention.}
    \label{fig:counterfactual}
\end{figure}

We intervene on emotion to test whether LEWM treats it as an active variable in future-state prediction. Holding the current state $\mathcal{S}_t$ and behavior trajectory $\mathcal{H}_t$ fixed, we replace $\mathcal{E}_t$ with each of the eight emotions and measure the cosine divergence from the reference prediction. As shown in Fig.~\ref{fig:counterfactual}, interventions under $do(\mathcal{E}_t{=}e')$ induce non-zero and structured changes across all three modalities, with high-arousal emotions (e.g., \textit{happiness}) producing the largest divergences and low-arousal emotions (e.g., \textit{neutral}) clustering tightly. These consistent, theory-aligned patterns indicate that LEWM actively conditions multimodal future-state prediction on emotion.

\section{Conclusion}
\label{sec:conclusion}

In this work, we introduce \textbf{EWH}, the first dataset for emotion-aware world-state transitions, and \textbf{LEWM}, a world model that treats emotion as a key variable in multimodal future-state prediction. EWH connects states, behaviors, and emotional changes, providing a benchmark for studying why human actions occur and how affective dynamics shape future states. Experiments on state prediction, emotion understanding, and general reasoning show that LEWM improves human-centered world modeling while preserving general capabilities. Together, EWH and LEWM establish a foundation for socially grounded agents that better model human behavior, interaction, and decision-making.

\bibliography{Bibliography-File}


\end{document}